\documentclass[letterpaper, 10 pt, conference]{ieeeconf}  

\IEEEoverridecommandlockouts                              

\usepackage{amssymb}  

\usepackage{subcaption}
\usepackage{color}
\usepackage{graphicx}
\usepackage{graphbox}
\usepackage{fixmath}
\usepackage{amsmath}
\usepackage{amssymb}

\usepackage{cleveref}
\usepackage{subcaption}
\usepackage{ifthen}
\usepackage{xspace}

\title{\LARGE \bf
Reducing Network Load via Message Utility Estimation for Decentralized Multirobot Teams\\
}

\author{Isabel M. Rayas~Fern\'andez$^{1}$, Christopher E. Denniston$^{1,\dagger}$, Gaurav S. Sukhatme$^{1,*}$
\thanks{This work was supported in part by the Southern California Coastal Water Research Project Authority under prime funding from the California State Water Resources Control Board on agreement number 19-003-150 and in part by USDA/NIFA award 2017-67007-26154.}
\thanks{This material is based upon work supported by the National Science Foundation Graduate Research Fellowship Program under Grant No. DGE-1842487. 
Any opinions, findings, and conclusions or recommendations expressed in this material are those of the author(s) and do not necessarily reflect the views of the National Science Foundation.
}
\thanks{$^{1}$Department of Computer Science, University of Southern California.}%
\thanks{$^{*}$G.S. Sukhatme holds concurrent appointments as a Professor at USC and as an Amazon Scholar. This paper describes work performed at USC and is not associated with Amazon.}%
\thanks{$^{\dagger}$C.E. Denniston is now at OffWorld, Inc. This paper describes work performed at USC and is not associated with OffWorld.}%
}

\newcommand{\gtlocations}{\mathbold{G^\#}}
\newcommand{\gtsensedlocations}{\mathbold{X^\#}}
\newcommand{\gtsensedvalues}{\mathbold{Y^\#}}

\newcommand{\sensedlocations}{\mathbold{X}}
\newcommand{\sensedvalues}{\mathbold{Y}}
\newcommand{\location}{g}
\newcommand{\sensedlocation}{x}

\newcommand{\nrobots}{N}
\newcommand{\spread}{\alpha}

\newcommand{\quantiles}{Q}

\newcommand{\estimatedquantilevalues}{\tilde{V}}

\newcommand{\estimatedquantilevaluesfinal}{\estimatedquantilevalues_{\textrm{final}}}
\newcommand{\quantilevalues}{V}

\newcommand{\ego}{i}
\newcommand{\other}{j}
\newcommand{\othermeasurements}{M^j}
\newcommand{\msg}{\mathcal{M}}
\newcommand{\utility}{u}
\newcommand{\aggutility}{U}
\newcommand{\utilthresh}{T}
\newcommand{\successprob}{p_{\textrm{success}}}
\newcommand{\newobjectivevalues}{\mathcal{Y}}

\newcommand{\objectivefunction}{f}
\newcommand{\argmax}{\arg\!\max}

\newcommand{\numtiles}{|\quantiles|}

\newcommand{\psig}{p \leq 5\textrm{e}{-2}} 

\newcommand{\wpres}[2]{T {=} #1,~p {\leq} #2}

\newif\ifshowrev
\newif\ifshowrevv

\begin{document}

\maketitle
\thispagestyle{empty}
\pagestyle{empty}

\begin{abstract}
We are motivated by quantile estimation of algae concentration in lakes
and how decentralized multirobot teams can effectively tackle this problem. 
We find that multirobot teams improve performance in this task over single robots, and communication-enabled teams further over communication-deprived teams;
however, real robots are resource-constrained, and communication networks cannot support arbitrary message loads, making na\"ive, constant information-sharing but also complex modeling and decision-making infeasible.
With this in mind, we propose online, locally computable metrics for determining the utility of transmitting a given message to the other team members and a decision-theoretic approach that chooses to transmit only the most useful messages, using a decentralized and independent framework for maintaining beliefs of other teammates.
We validate our approach in simulation on a real-world aquatic dataset, 
and we show that restricting communication via a utility estimation method based on the expected impact of a message on future teammate behavior results in a 42\% decrease in network load while simultaneously decreasing quantile estimation error by 1.84\%.
\vspace{-.07in}
\end{abstract}

\section{Introduction}
\label{sec:intro}

\begin{figure}[t]
    \centering
    \includegraphics[width=\columnwidth]{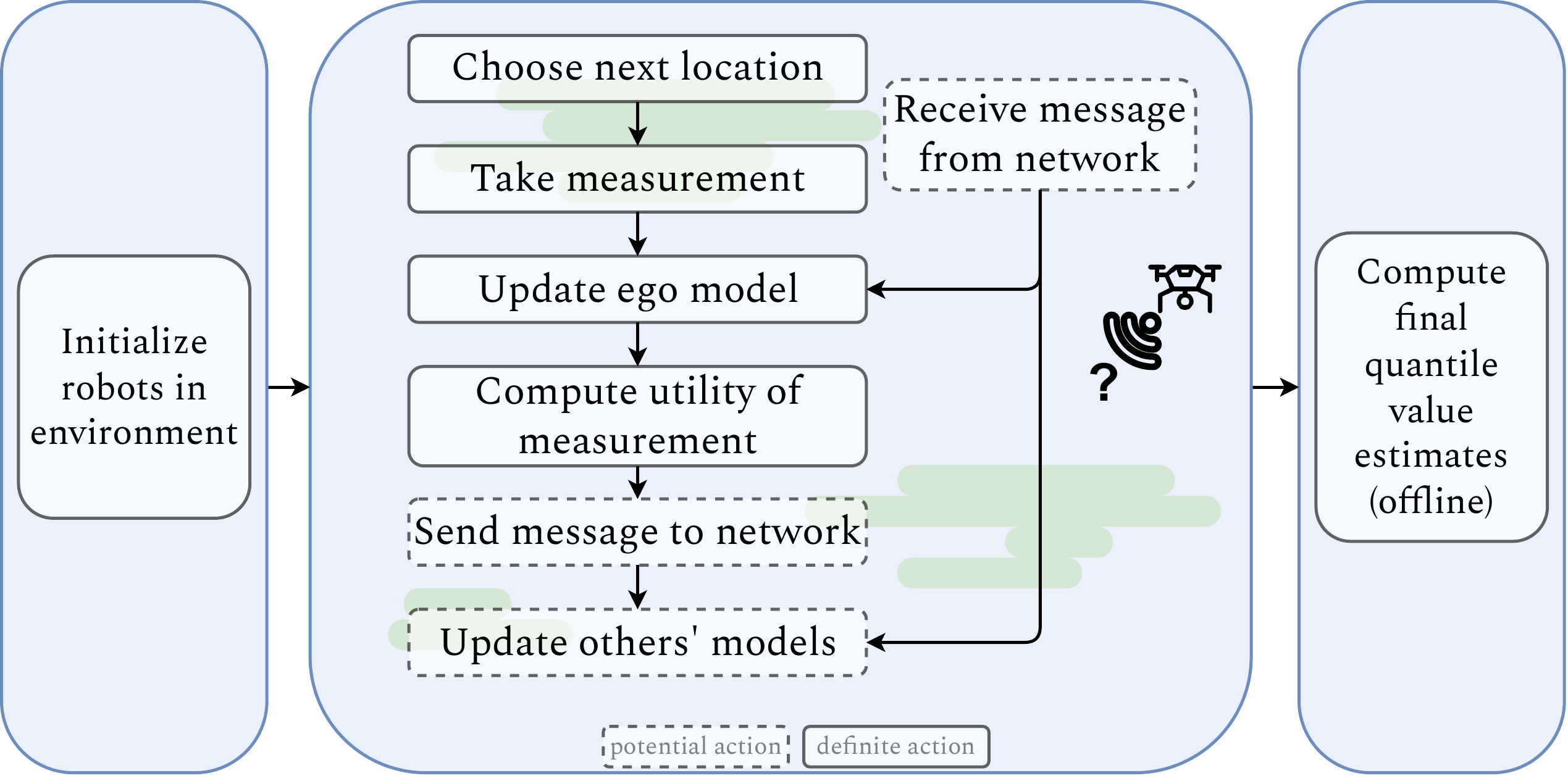}
    \caption{Overview of problem pipeline. 
    The approach consists of 3 phases: Initialization, Exploration, and Aggregation. 
    The bulk of the work happens during the decentralized exploration phase, where robots repeatedly choose the next location to visit, measure the environment, optionally broadcast the measurement in a message, and update their models until their budget $B_T$ is exhausted.}
    \label{fig:comms-hero}
    \vspace{-0.2in}
\end{figure}

Multirobot teams have shown promise for effectively exploring and accurately estimating quantile values to aid understanding of natural environments \cite{rayas2023study}.
When studying spatially varying fields, quantile estimation is useful for describing varied spreads of interest, and entails using measurements from the environment to estimate the values of desired quantiles (e.g., the 25th percentile is equivalent to the 0.25 quantile) \cite{rayas2022informative}.
We are interested in using a team of communication-enabled mobile robots to collaboratively explore a natural environment such that we achieve an accurate estimate of quantile values in that environment.
However, realistic multirobot deployments must consider limitations in communication capabilities; here, we focus on reducing the network load as measured by the number of attempted transmitted messages with the goal of sacrificing minimal quantile estimation accuracy. 
We address the question of how to decide if a message is worth sending.
Specifically, we:
\begin{itemize}
    \item Propose methods to reduce network load for bandwidth- and resource-constrained decentralized multirobot teams that leverage utility-based message evaluation, with no guarantees on information from teammates;
    \item Describe a decision-theoretic approach to choosing whether to transmit a message to the network rooted in the utility-based evaluation and a cost of transmission;
    \item Present results showing significant reductions in network load are achievable with minor corresponding drops in performance on the problem task of estimating quantile values, even with probabilistic transmission failure and a distance-based transmission probability, on a dataset from a real aquatic environment;
    \item Discuss results in the context of real-world applications.
\end{itemize}

We assume a decentralized team where each robot plans independently and is not guaranteed to receive information from  others.
However, we assume a communication protocol such that a robot can, at each step, decide to broadcast its newly acquired measurements to the network. 
The network has limited bandwidth, so robots are incentivized to only send those that will be most beneficial to the others.
We encode the cost of loading the network with a message through a threshold value, where utilities higher than the threshold have a higher benefit than cost.
With these assumptions in place, the question of how to evaluate a potential message for its utility emerges as a key consideration to this problem; in this work, we propose several approaches to solve it.
A general overview can be seen in \Cref{fig:comms-hero}.

\section{Related Work}
There is an extensive literature in utility-based communication, and our methods share similarities with previous work. Work that has considered the problem
of deciding what or when to communicate is largely motivated by the same desire to respect bandwidth limitations that
real networks must face and to reduce communication costs.
However, particular methods differ in how they approach
 solutions and in what context the problem is situated.
Some of the varied contexts in which utility-based decision-making for inter-robot communication has been investigated
include months-long data collection to be routed to a base station~\cite{padhy2010utility}, disaster response \cite{lieser2019understanding}, non-cooperative multi-agent path finding~\cite{ma2021learning}, and
those with request-reply protocols~\cite{ma2021learning,zhang2019efficient}.

One common approach has been to use message category, type, or intrinsic data features to determine which messages are most valuable to send \cite{padhy2010utility, lieser2019understanding, barcis2021information}.
These approaches typically leverage domain knowledge for their particular application.
Previous work aimed at the problem of multirobot localization has applied directed acyclic graphs built by forward-simulating the robot team under both transmission and non-transmission of each message \cite{anderson2021communication}. 
The goal then is to reach the best group localization accuracy while minimizing the cost (number of communications required in the DAG).

The expected effect of the message on robot actions has also been used to determine utility \cite{ma2021learning, zhang2019efficient}. 
One approach for restricting the number of communications when robots are tasked with selecting goal waypoints to visit is to use a heuristic based on limiting the new path's length, and only request information from others if the path would be within a certain factor length of the current path \cite{best2020decentralised}.
Monte Carlo Tree Search (MCTS) has been used within a general middleware in order to make decisions on whether a message should be sent to the network \cite{barcis2021information}. Their approach takes into account the temporal aspect of message utility potentially changing over time, and the tree built at each decision point includes all previous and future messages. 
However, it does not explicitly encode network constraints and assumes that robots are always within communication range of each other.

Another class of solutions propose metrics to determine utility based on the difference in expected reward after incorporating a message \cite{simonetto2014distributed,
marcotte2020optimizing, unhelkar2016contact}. 
In \cite{velagapudi2007maintaining}, the authors proposed a utility function computed as the sum of two competing factors: maximizing utility gained by the entire team and minimizing the number of times a message is sent. 
There, the network is point-to-point, and thus at each step, a decision is made whether to forward the received message to a random recipient or to cease forwarding.
[7] and [8] also include a
measure of utility based on expected reward, and the teams
are decentralized in terms of planning and reward calculation.
A major difference between these works and ours is that our end goal is to explore the
environment in whatever way results in the best quantile
estimates, while those define a static goal location
for the robots to reach. We also exclusively choose to
plan using a greedy algorithm for computational efficiency
and tractability, motivated by resource-constrained systems,
and do not forward-simulate the models of other robots.

Significantly, we make the (more realistic) assumption
of stochastic communication on top of the communication
decision; just because a robot decides to send a message
does not guarantee reception by all or any of the recipients.
We do not assume perfect measurement sensing. Despite
our relatively simple planning and modeling methods and
more failure-prone communication assumptions, we are able
to achieve a significant decrease in network loading while
maintaining good task performance.

\section{Methods}
\label{sec:methods}

Our formulation extends the setup in~\cite{rayas2023study} to which we refer the reader for details; we briefly mention relevant definitions here. 
The team consists of $\nrobots$ aerial robots, each operating with a fixed planning budget $B_T$ at a fixed height in a discretized environment
$\gtlocations \subset
\mathbb{R}^2$. 
Robots can take one of four possible actions at each step: $\pm x$ or $\pm y$.
Since our focus is on inter-robot communications, we assume a low-level controller, a collision avoidance routine, and perfect localization.
The human operator supplies a set of arbitrary quantiles of interest $\quantiles$, and the end goal is to produce accurate estimates $\estimatedquantilevalues$ of $\quantiles$ where the true values are
defined as $\quantilevalues = \textit{quantiles}(\gtsensedvalues,\quantiles)$, 
the set of all possible locations a robot can measure with its sensor are $\gtsensedlocations$, and 
 the corresponding true values at those locations are $\gtsensedvalues$.
Each robot has a camera with which it can take noisy measurements $\sensedvalues$ representing the concentration of algae at those pixel locations in the environment $\sensedlocations$.
Throughout the deployment, each robot collects measurements $\sensedlocations, \sensedvalues$ both firsthand via its camera as well as via any messages received from others through the network.
The robots use all their collected measurements to construct and update their internal Gaussian process (GP) model of the environment.
%
Robot $\ego$'s estimate of the quantile values 
is given by
\begin{equation}
    \estimatedquantilevalues_{\ego} = quantiles(\mu_{GP^{\ego}}(\gtsensedlocations),\quantiles)
\end{equation}
where $\mu_{GP^{\ego}}(\gtsensedlocations)$ is the estimate of all possible locations
using the GP conditioned on the measurements $\sensedlocations^{\ego}$ and $\sensedvalues^{\ego}$ that $\ego$ has collected thus far.

In what follows, when viewing the problem from the perspective of a given robot, we will refer to it as $\ego$ or the ego robot and to any of the other $\nrobots$ robots as $\other$. 
Variables superscripted with $\other$ such as $GP^{\other}$ indicate the GP for $\other$ \textit{from the perspective of $\ego$}.

Motivated by the computational constraints of resource-constrained systems, we opt to use a locally computable, online, greedy planning policy.
At each planning step $t$, $\ego$ maximizes an objective function $\objectivefunction$ over feasible next measurement locations.
We use a modified version of the quantile standard error objective function
to evaluate a proposed location for time $t$
\cite{rayas2022informative},
which measures the difference in the standard error of the estimated quantile values using 
two versions of the robot model: the current model $\mu_{GP^{\ego}_{t-1}}$, and the model updated with the hypothetical new values at the proposed locations $\mu_{GP^{\ego}_{t}}$.
We modify $\objectivefunction$ such that it is conditioned not only on the proposed measurement locations $\sensedlocations$, but also the values associated with the locations (denoted by $\newobjectivevalues$ generally) and the GP to be used.
The reasoning is that $\objectivefunction$ is also used later in the process for computing message utilities, and this allows it to generalize to these cases when $\newobjectivevalues$ and $GP$ may take on different values.
When $\objectivefunction$ is used in the planning step by $\ego$ to select $\ego$'s next location, $\newobjectivevalues$ are the expected values at the proposed location $\sensedlocations$: $\mu_{GP^{\ego}_{t-1}}(\sensedlocations)$ (just like in \cite{rayas2022informative}).
However, when $\objectivefunction$ is used by $\ego$ to determine the utility of $\other$ receiving a real measurement that $\ego$ has already acquired, $\newobjectivevalues$ are those acquired $\sensedvalues$, and the GP may be either $GP^{\ego}$ or $GP^{\other}$ (see \Cref{ssec:modeling}).
The final term encourages exploration of high-variance areas.
\begin{equation}
\label{eq:objective}
    \objectivefunction(\sensedlocations_i; \newobjectivevalues, GP) =
\frac{d}{\numtiles} +
\sum_{\sensedlocation_j \in \sensedlocations_i} c\sigma^2(\sensedlocation_j)
\end{equation}
\begin{equation}
    d =
\|se(\mu_{GP_{t-1}}(\gtsensedlocations),\quantiles) -
se(\mu_{GP_{t}}(\gtsensedlocations),\quantiles) \|_{1}
\end{equation}

\subsection{Summary of Process}
\label{ssec:procedure}
The approach consists of 3 phases: Initialization, Exploration, and Aggregation.
These are described at a high level here; further details are in the sections following.

\subsubsection{Initialization}
\label{sssec:procedure-init}
Robots begin distributed in the environment according to the initial location spread $\spread$ (\cite{rayas2023study} provides details). 
We assume each robot is given the number of other robots and their corresponding starting locations.

\subsubsection{Exploration}
\label{sssec:procedure-explore}
The bulk of the work happens during the decentralized exploration phase, where robots repeatedly choose the next location to visit, measure the environment, optionally broadcast the measurement in a message, and update their models until their budget $B_T$ is exhausted. The next measurement locations at $t$ are selected according to
\begin{equation}
\label{eq:greedy-planning}
    \sensedlocations^{\ego}_t = \argmax_{\sensedlocations \subset \sensedlocations^{\ego}_{\textrm{next}}} \objectivefunction (\sensedlocations;  \mu_{GP^{\ego}}(\sensedlocations), GP^{\ego})
\end{equation}
where $\sensedlocations_{\textrm{next}}$ contains the sets of locations that could be sensed one step from the current location, $\location^{\ego}_t$.
On top of this planning cycle, $\ego$ may receive a message from another $\other$ at any point, leading to further model updates (\Cref{ssec:modeling}).

\subsubsection{Aggregation}
\label{sssec:procedure-aggregate}
Once the robots have reached the allotted budgets, their data is collected. The operator can produce an aggregate GP model of the environment offline using it and, with that, obtain final estimates of quantile values:
\begin{equation}
\label{eq:final-qest}
    \estimatedquantilevaluesfinal = quantiles(\mu_{GP_\textrm{aggregate}}(\gtsensedlocations),\quantiles)
\end{equation}

\subsection{Communication Framework}
\label{ssec:comms}
The communication network consists of $\nrobots$ robots, each with the ability to broadcast a message at each timestep. 
A message $\msg_t^{\ego} = (\ego, \location^{\ego}_t, \sensedlocations^{\ego}_t, \sensedvalues^{\ego}_t)$ sent by $\ego$ at time $t$ is composed of a header containing the sending robot's ID number and current location, and a body containing the current set of measurements. 
A sent message is (independently) transmitted successfully to every other robot $\other$ in the network subject to the probability 
depending on distance and parameterized by the dropoff rate $\eta$ and communication radius $r$ \cite{rayas2023study}:
\begin{equation}
\label{eq:comms-probability}
    \successprob(distance) = \frac{1}{1 + e^{\eta (distance - r)}}
\end{equation}
A successfully transmitted message will be received whole, uncorrupted, and without delay. 
If a message is unsuccessfully transmitted to $\other$, it is considered lost for that robot.
In general, we assume $\ego$ does not receive a handshake or confirmation of whether its message was received successfully or not by $\other$.
We assume that robots can self-localize and that, within the radius $r$, they can accurately sense $\location^{\other}$, which they do for each $\other$ at each step. 

\subsection{Modeling Others and Updating Models}
\label{ssec:modeling}
A central consideration is how to model other robots using incomplete information since such a model is necessary to adequately assess the utility that any given message will have for others. 
The ego robot $\ego$ maintains a separate model $(\location^{\other},~ GP^{\other})$ for each $\other$. 
$\location^{\other}$ is the most up-to-date location $\ego$ has, and is updated in two cases.
First, by $\ego$'s sensing capabilities at each step if $\other$ is within radius $r$, 
and second, anytime $\ego$ receives a message from $\other$, via the header.
$GP^{\other}$ is a GP based on the measurements $\ego$ believes $\other$ has.
We denote this as $\othermeasurements := (\sensedlocations^{\other}, \sensedvalues^{\other})$.
%
$\othermeasurements$, and $GP^{\other}$, which is defined by it, can also be updated in two scenarios.
The first is when $\ego$ receives a message $\msg$ from $\other$, since it is clear that $\other$ must possess the data contained.
The second is more uncertain. If $\ego$ broadcasts $\msg$ to the network, it does not know whether it was successfully transmitted to any $\other$; however, recall that $\ego$ can sense the location of $\other$ within $r$. We use this to determine a proxy for reasonable confidence of transmission success: If $\other$ is within $r$, $\ego$ believes $\msg$ was received and thus updates $\othermeasurements$.%

\subsection{Computing Message Utility}
\label{ssec:utility}
We introduce several methods for computing the utility $\utility^{\other}$ of $\msg^{\ego} = (\ego, \location^{\ego}, \sensedlocations, \sensedvalues)$ (dropping the super/subscripts for readability). 
The first,
\textit{Reward}, defines $\utility_{rw}$ as the reward $\ego$ believes $\other$ would receive, using the objective function $\objectivefunction$:
\begin{equation}
    \utility_{rw}(\sensedlocations, \sensedvalues, \other) = 
    \begin{cases}
      \infty, & \text{if} \quad | \sensedlocations^{\other} | = 0\\
      \objectivefunction(\sensedlocations ; \sensedvalues, GP^{\other}), & \text{otherwise}
    \end{cases}
\end{equation}

We call the second method \textit{Action}. Intuitively, it considers a message utile if $\ego$ believes that $\other$ receiving those measurements would result in $\other$ taking a different action than it would without them. 
Then $\utility_{ac}$ is defined as $\infty$ if the actions $\sensedlocations_{\textrm{next,w/o}}$ and $\sensedlocations_{\textrm{next,w/}}$ differ as computed using \Cref{eq:greedy-planning}, and if they do not, $\utility_{ac}$ takes the value of $\utility_{rw}$:
\begin{equation}
    \utility_{ac}(\sensedlocations, \sensedvalues, \other) = 
    \begin{cases}
      \infty, & \text{if} ~ | \sensedlocations^{\other} | = 0\\
      \infty, & \text{if} ~ \sensedlocations_{\textrm{next,w/o}} \neq \sensedlocations_{\textrm{next,w/}} \\
      \utility_{rw}(\sensedlocations, \sensedvalues , \other), & \text{otherwise}
    \end{cases}
\end{equation}
Both $\utility_{rw}$ and $\utility_{ac}$ include the caveat that, if $\ego$ believes $\other$ has collected no measurements yet, the message should be sent.

The last three methods presented are simpler in that they do not make use of the model of $\other$.
In the third method, \textit{Ego-reward}, $\ego$ uses the reward that it itself actually received from its measurement, and directly sets that as the utility for all other $\other$:
\begin{equation}
    \utility_{ego}(\sensedlocations, \sensedvalues, \other) = 
     \objectivefunction(\sensedlocations, \sensedvalues ; GP^{\ego})
\end{equation}
Finally, \textit{Always} and \textit{Never} are constant baselines, which always assign either $\infty$ or $0$:
\begin{equation}
    \utility_{al}(\sensedlocations, \sensedvalues, \other) = 
     \infty
\end{equation}
\begin{equation}
    \utility_{nv}(\sensedlocations, \sensedvalues, \other) = 
     0
\end{equation}

\subsection{Deciding to Communicate}
With the utilities of a message $\{\utility^{\other}\}$ computed for each $\other$, $\ego$ now faces the problem of deciding whether the utility overcomes the cost of loading the network with a message. 
Since the communication is broadcast-based rather than point-to-point, $\ego$ must first compute a single utility value $\aggutility$ for the team and then determine if it crosses a utility threshold $\utilthresh$ representing the cost.
For this, we describe a method 
for aggregating a set of utilities 
which computes 
\textit{expected utility}, weighting utilities by the estimated transmission success probability:
\begin{equation}
\label{eq:expected-utility}
    \aggutility_{EU}(\{\utility^{\other}\}) = \frac{1}{\nrobots - 1} \sum_{\other} (p_{est}(\other) \cdot \utility^{\other})
\end{equation}
\begin{equation}
\label{eq:estimated-success-probability}
    p_{est}(\other) = 
    \begin{cases}
      \successprob(d^{\other}), & \text{if} \quad d^{\other} \leq r \\
      \successprob(r), & \text{otherwise}
    \end{cases}
\end{equation}
where $d^{\other} = ||\location^{\ego} - \location^{\other}||_2$ and $r$ is as defined in \Cref{ssec:comms}.
The estimated success probability, computed in \Cref{eq:estimated-success-probability}, directly uses the probability based on the distance to $\other$ if it is known (i.e., if $\other$ is within the sensing radius $r$), but defaults to an optimistic probability estimate for any $\other$ that is beyond the sensing radius by using $r$ as the assumed distance.
This optimistic heuristic prevents overly restrictive utility values in the face of unknown teammate positions.
%
Note that 
if any $\utility^{\other}$ is $\infty$, then $\aggutility$ is considered past the threshold $\utilthresh$.
If $\aggutility \geq \utilthresh$, $\ego$ decides to transmit $\msg$.%


\begin{table*}
\centering
\caption{
  Number of attempted and successful transmitted messages per utility method. 
  Restricting transmissions with Action results in more than $40\%$ reduction in network load (attempted) while improving mean final quantile estimation error  by nearly $2\%$ and only resulting in a less than $25\%$ decrease in successful transmissions.
  Results reported from $60$ experiments total of length 40 steps.
  All experiments use $\nrobots=4$, so 
  the most messages possibly sent total is $40\times(\nrobots-1)=120$.
  }
  
  \begin{tabular}{c||cc|cc|c}
    \textit{Method} & \textit{Attempted} & \textit{Median} & \textit{Successful} & \textit{Median} & \textit{Median} \\ 
     & \textit{($\mu, \sigma$)} & \textit{Decrease} & \textit{($\mu, \sigma$)} &  \textit{Decrease} &  \textit{RMSE Increase} \\ 
    \hline 
    %
%
%
Action & (66.6, 27.95) & 42.5\% & (38.2, 23.85) & 24.32\% & -1.84\% \\
Always & (120.0, 0.0) & -- & (44.93, 16.78) & -- & -- \\
Ego-reward & (76.2, 28.33) & 27.5\% & (34.4, 20.27) & 32.2\% & 5.96\% \\
Never & (0.0, 0.0) & 100.0\% & (0.0, 0.0) & 100.0 & 73.98\% \\
Reward & (68.2, 32.1) & 52.5\% & (36.0, 22.77) & 35.96\% & 16.03\% \\
  \end{tabular}
  \label{tab:comms}
  \vspace{-0.2in}
\end{table*}

\begin{figure}
  \centering
 \includegraphics[width=\columnwidth]{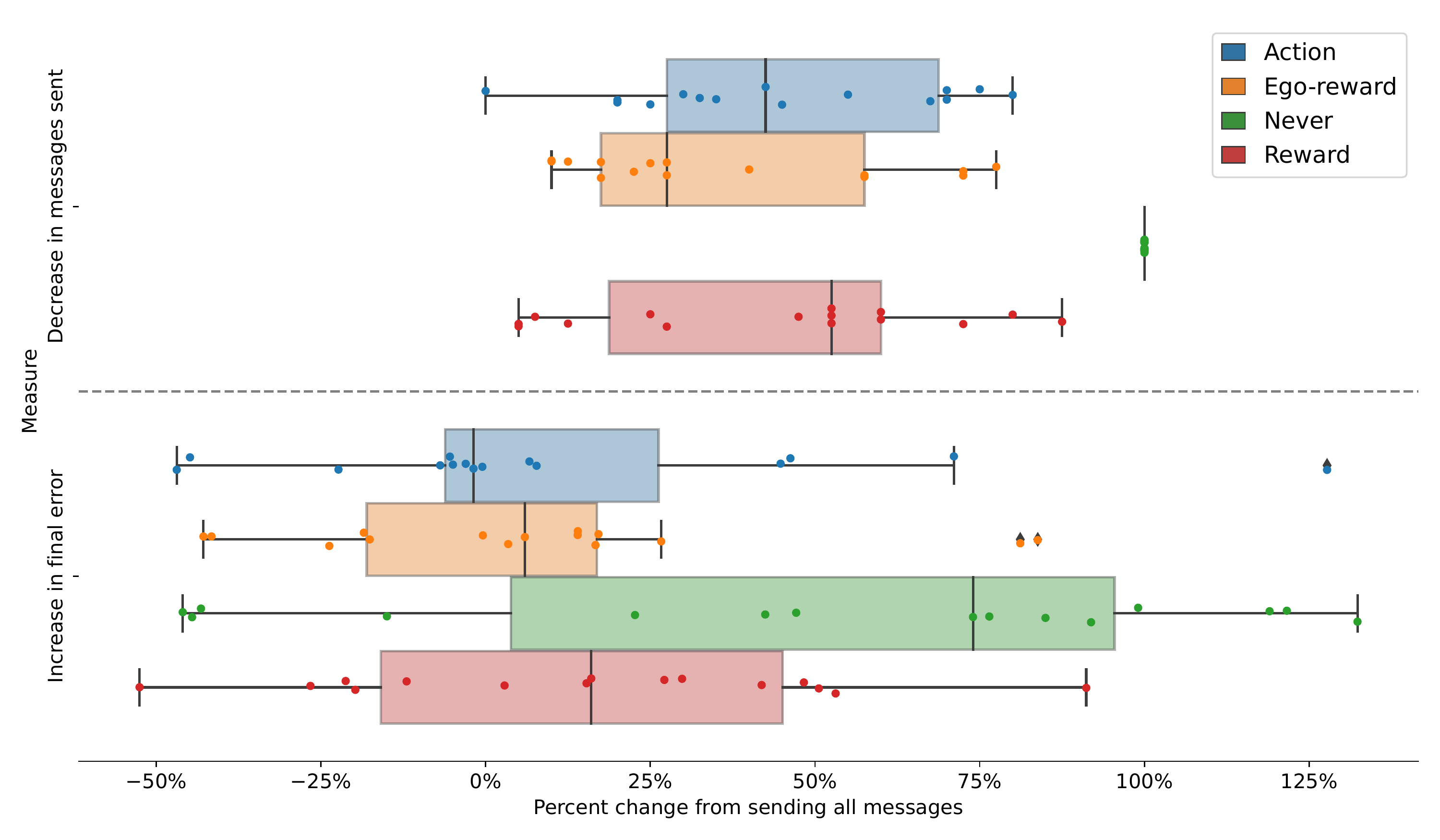}
  \caption{
  Tradeoff in network load vs. task performance for different utility methods, compared to always sending messages.
  Top: Percent decrease in messages sent (network load). Higher is better.
  Bottom: Percent increase in final quantile estimation error. Lower is better. 
  }
  \label{fig:comms-tradeoff}
\end{figure}

\begin{figure}
  \centering
 \includegraphics[width=\columnwidth]{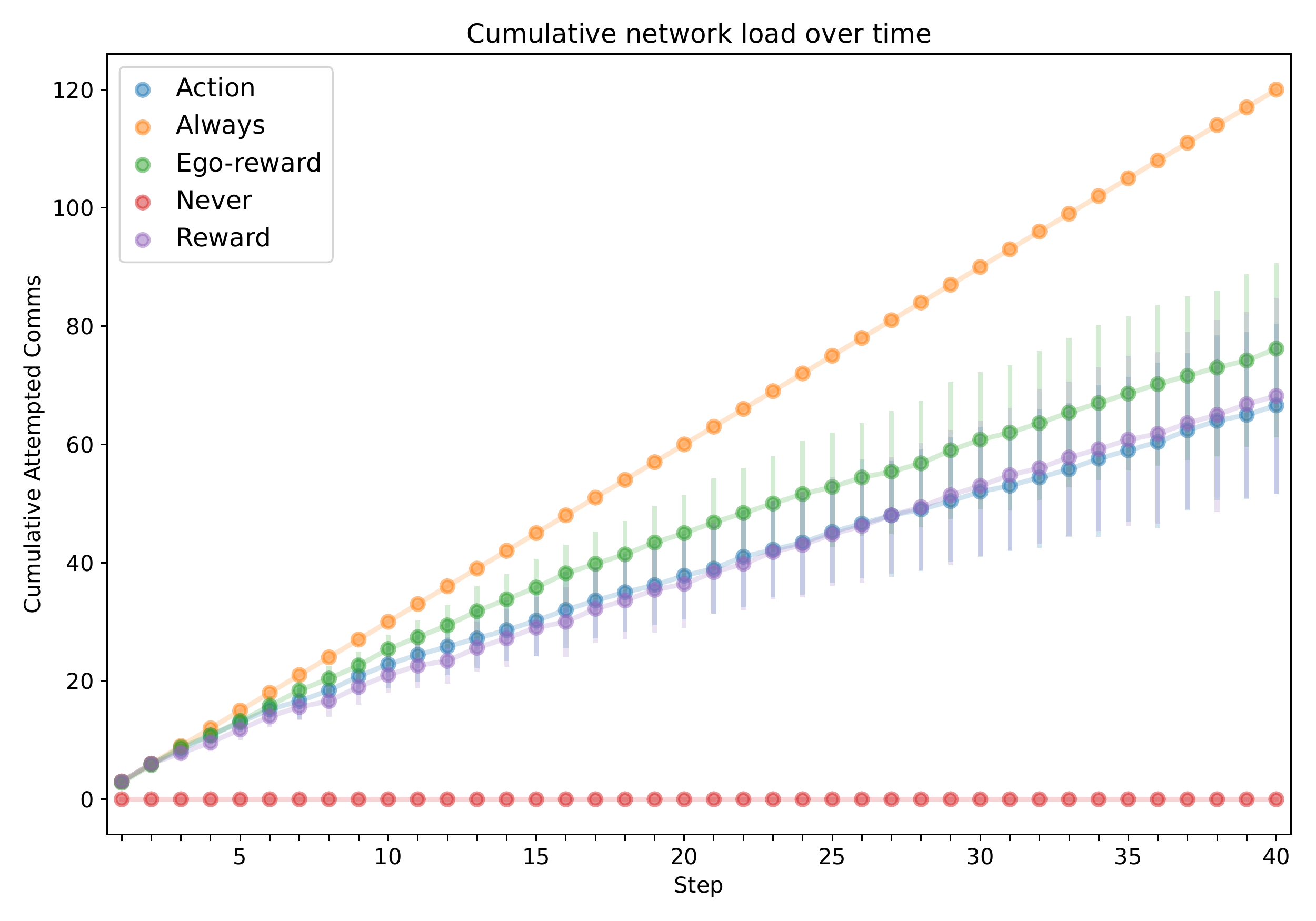}
  \caption{
  Cumulative network load, as measured by messages sent, over time. 
  Each step on the x-axis represents a robot taking a planning step. 
  }
  \label{fig:comms-cumulative}
  \vspace{-0.2in}
\end{figure}

\begin{figure}
  \centering
 \includegraphics[width=\columnwidth]{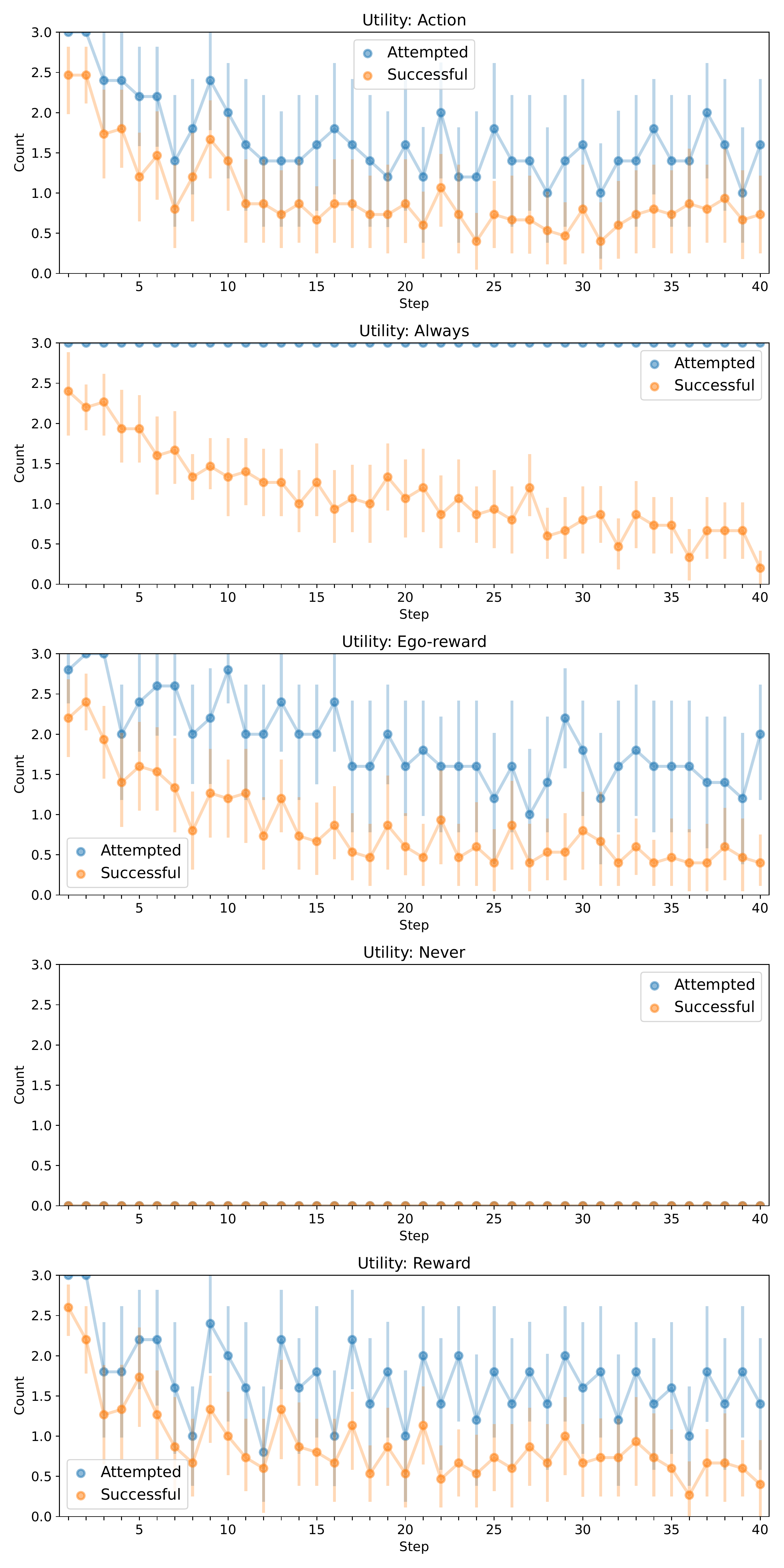}
  \caption{
  Amount of attempted and successful transmissions over time for different utility methods.
  Each step on the x-axis represents a robot taking a planning step. 
  }
  \label{fig:comms}
\end{figure}

\section{Experiments and Discussion}

We now present experimental results for the methods described in \Cref{sec:methods}.
%
We test performance on 3 sets of quantiles $\quantiles$: Quartiles (0.25, 0.5, 0.75), Median-Extrema (0.5, 0.9, 0.99), and Extrema (0.9, 0.95, 0.99).
For each combination of parameters, we run 5 seeds, resulting in 60 total instances.
We fix the following parameters: $\nrobots = 4$; budget $B_T = 10$ ($40$ steps total); initial spread $\spread = 0.2$; $\eta = 0.4$; communication radius $r = 15.0$; thresholds $\utilthresh_{rw} = 2.8 \times 10^{-4}$ and $\utilthresh_{ego} = 8.3 \times 10^{-5}$.
$\utilthresh_{rw}$ is used for Reward and Action, while $\utilthresh_{ego}$ is used for Ego-reward.
We selected these $\utilthresh$ experimentally by observing the received $\utility$ values during testing and setting $\utilthresh$ to the approximate 25th percentile of observed rewards. 
This can be interpreted as only messages that are in the top 75\% of utility being worth the cost of sending to the network, while those in the lower quarter are not.

The robots operate in simulation using a real-world dataset collected with a hyperspectral camera mounted on a drone from a lake in California. The environment is roughly $80 \times 60$ meters, discretized to $\gtlocations = 25 \times 25$ locations. Each set of measurements is an image of $5 \times 5$ pixel intensities for the 400nm channel of the hyperspectral image, normalized to $[0,1]$. To each measurement, we add zero-mean Gaussian noise with a standard deviation $0.05$.%

\subsection{Utility Methods Comparison}
\begin{figure}
  \centering
         \includegraphics[width=0.9\columnwidth]{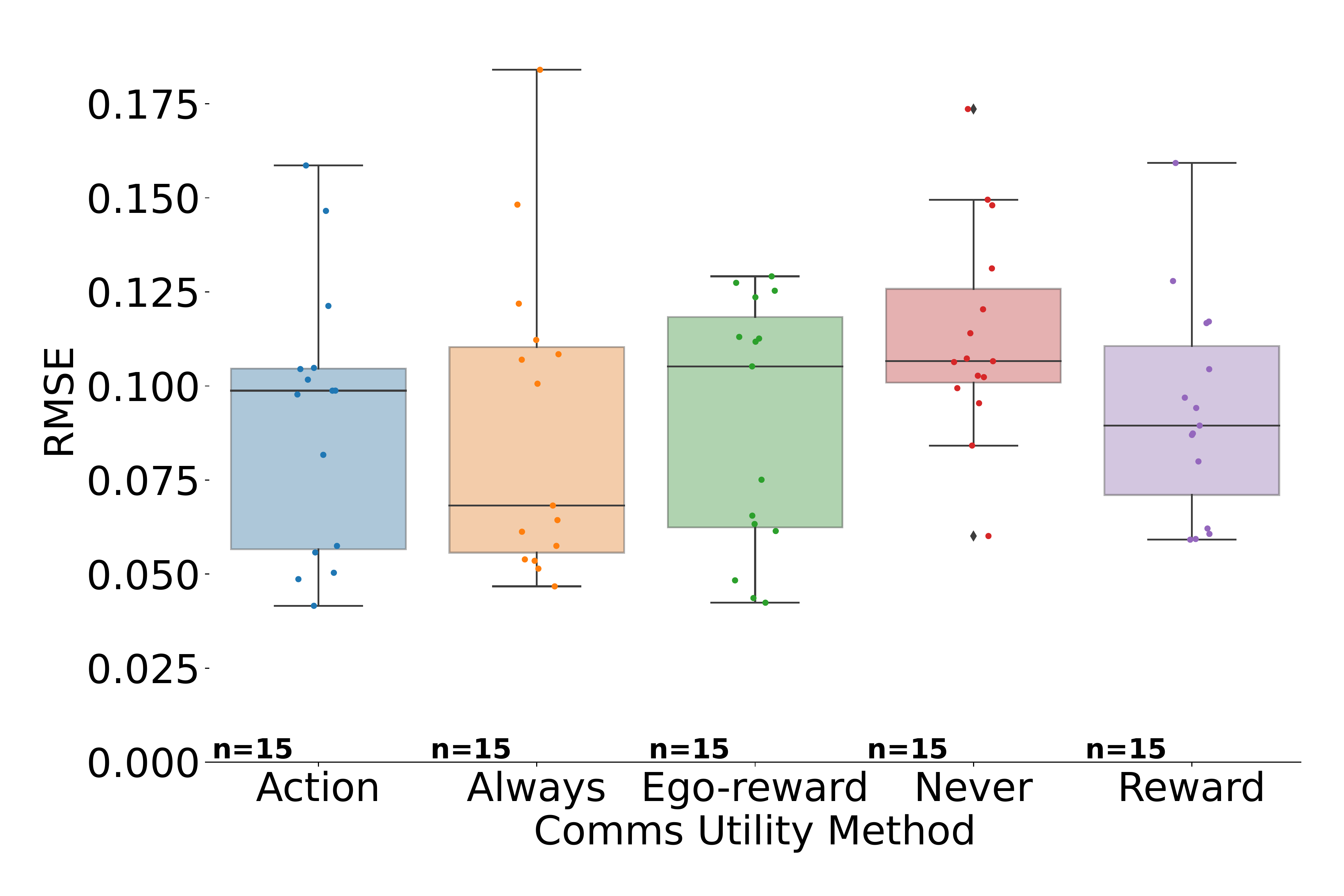}
  \caption{
  Final estimation error for different utility methods. 
  }
  \label{fig:rmse-utility}
  \vspace{-0.1in}
  \end{figure}

\begin{figure}
  \centering
  %
 \includegraphics[trim={0.5cm 3cm 0.5cm 3.5cm}, clip, width=0.8\columnwidth]{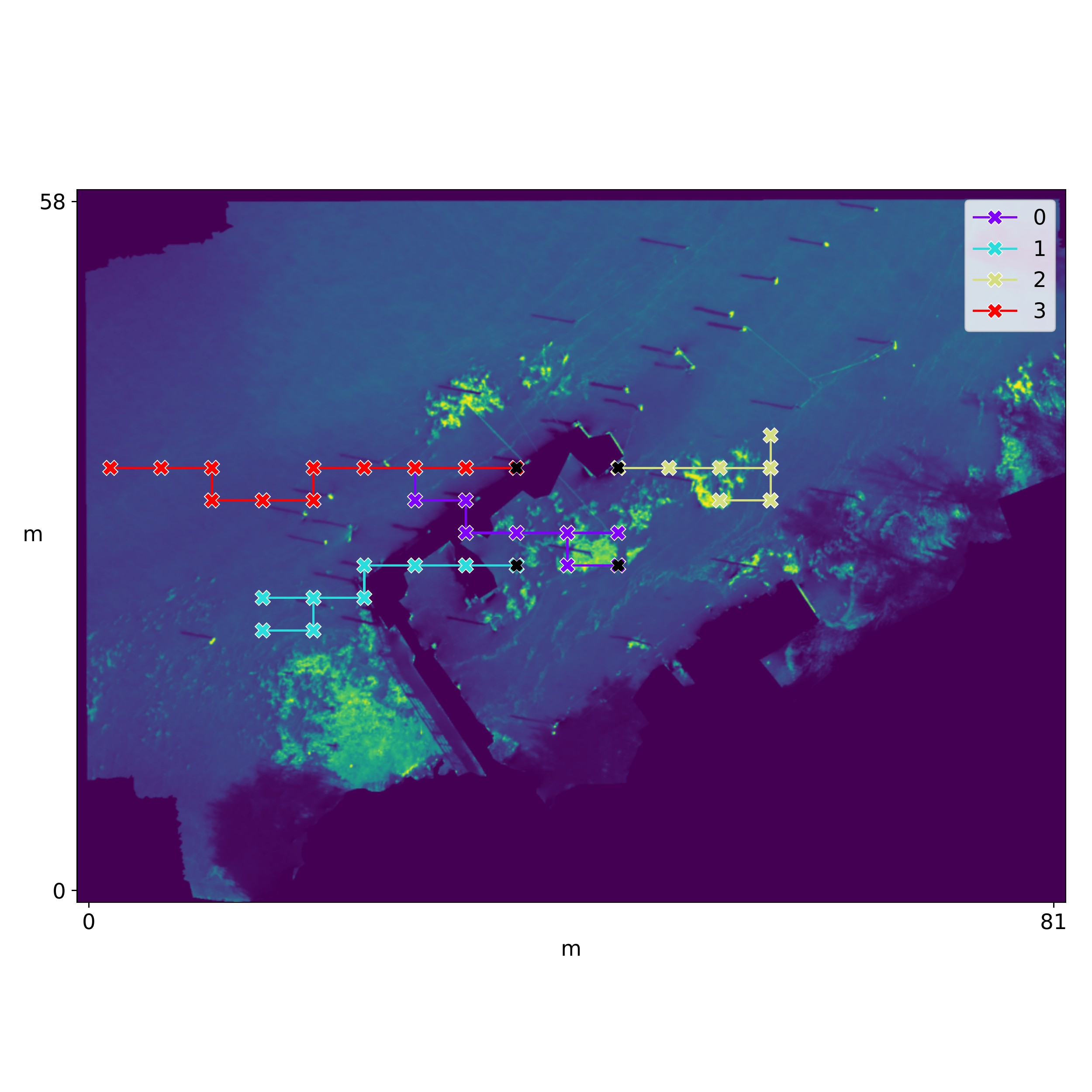}
  %
 \includegraphics[trim={0.5cm 3cm 0.5cm 3.5cm}, clip, width=0.8\columnwidth]{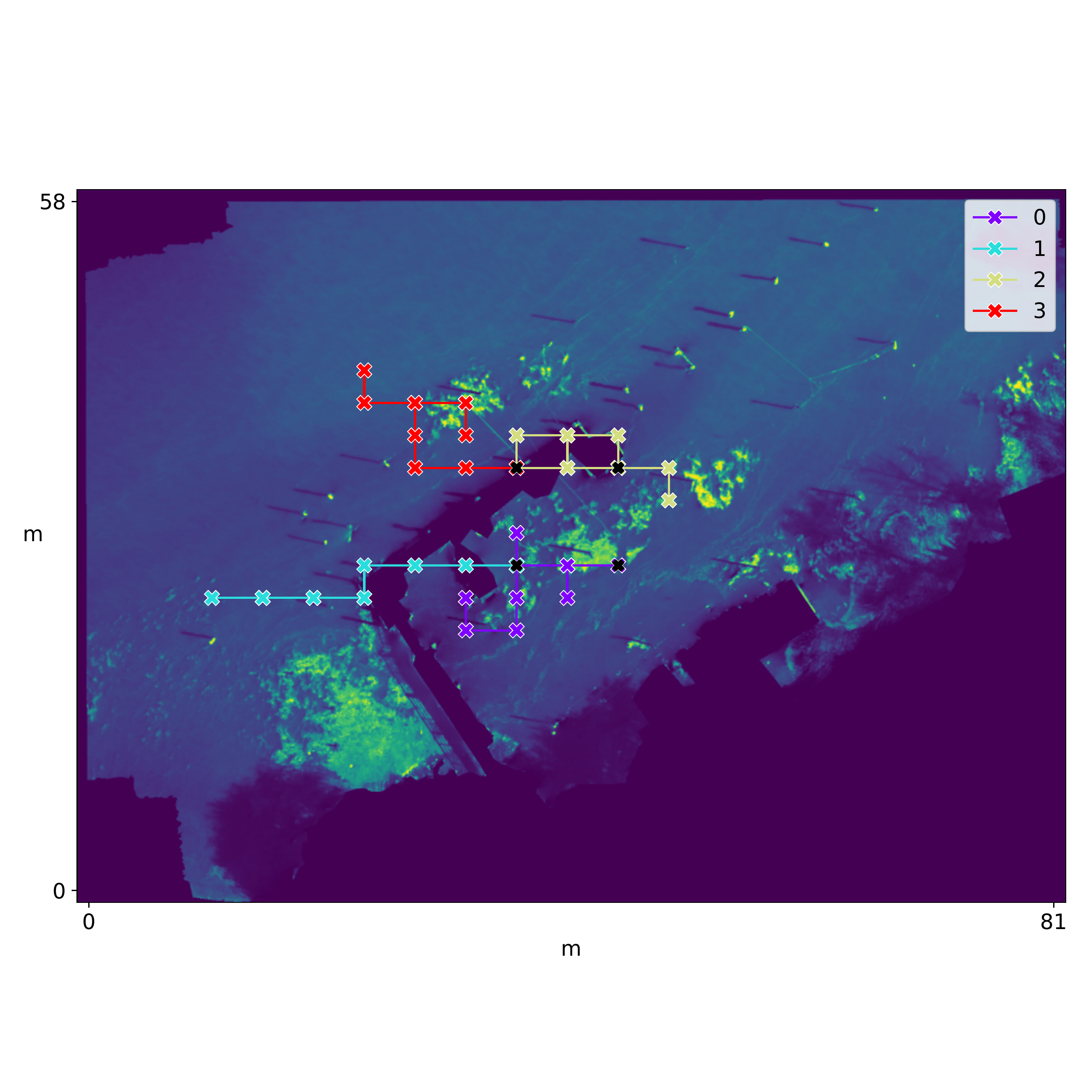}
  \caption{
  Example paths using two different utility methods with $\quantiles$ = $(0.9, 0.95, 0.99)$.
  Top: Action. Bottom: Reward.
}
  \label{fig:rasters}
  \vspace{-0.2in}
\end{figure}

We investigate the effectiveness of the different methods on performance in terms of both the network load as well as the final error (RMSE) between the estimated quantile values $\estimatedquantilevaluesfinal$ and $\quantilevalues$.
Units are pixel intensity normalized to the range $[0,1]$.
\Cref{tab:comms} displays numerical results, including a comparison of each method to Always in terms of the median percent change of each reported metric. 
These results are shown graphically in \Cref{fig:comms-tradeoff}, illustrating the tradeoff between network load and task performance.
\Cref{fig:comms-cumulative} shows the cumulative network load as measured by the attempted transmissions over time, while 
\Cref{fig:comms} shows the number of attempted (blue) and successful (orange) message transmissions plotted over time.
\Cref{fig:rmse-utility} shows boxplots of the final RMSE for different utility methods.
\Cref{fig:rasters} illustrates example paths taken 
superimposed on the hyperspectral image of the lake.

In \Cref{fig:comms}, we observe that 
across all methods, successful transmissions generally decrease over time due to robots spreading out.
Comparing Never to the other methods in \Cref{fig:rmse-utility}, we can confirm our previous findings that inter-robot communication improves performance.
However, we are mainly interested in the effect of decreasing network load.

When we analyze results in \Cref{tab:comms} and \Cref{fig:comms-tradeoff}, we see that severe reductions in communication do not necessarily result in severe decreases in performance. 
We see that Action results in the best performance: 
Compared to Always, there is more than $40\%$ decrease in network load as measured by attempted message transmissions, but this translates to less than $25\%$ decrease in actual (successful) message transmissions. 
Most notably, in terms of final error, there is a $1.84\%$ decrease in mean RMSE, indicating that performance, on average, improves slightly even with restricted communication.

Other reward-based methods, Reward and Ego-reward, also perform favorably.
\Cref{fig:comms,fig:comms-cumulative} suggest that all three proposed methods have message transmission rates that generally decrease over time.
We speculate that this may be due to two reasons.
First, as robots explore more of the environment and receive more information from the others, their beliefs of their teammates will grow such that $|\sensedlocations^{\other}|$ increases, leading to more accurate environment models and thus less value in receiving an additional set of measurements.
Second, with more exploration time, robots are more likely to be situated at further distances from each other. With distance, the expected utility will decrease according to the transmission probability, and even messages believed to be useful may not be deemed worthy because of their unlikelihood of successful delivery. 
We note, however, that the optimistic assumption in \Cref{eq:estimated-success-probability} will prevent a highly useful message from not being sent due to distance.

As shown qualitatively in \Cref{fig:comms,fig:comms-cumulative} and quantitatively in \Cref{tab:comms}, Action, Reward, and Ego-reward all have lower network load than Always, but their relative loads vary as do their final estimation error.
In particular, Reward results in the highest decrease in network load (excluding Never) at $52.5\%$, but also the largest increase in final error of more than $16\%$. 
Thus, such a metric may be valuable when network bandwidth is severely limited.
Ego-reward, on the other hand, has the smallest decrease in network load at $27.5\%$, but a relatively small increase in error at less than $6\%$. 
Ego-reward has the benefit of not requiring any onboard modeling of teammate beliefs; thus, this method could be powerful when onboard compute is limited. 
Action is the overall best performer, with a decrease in network load of $42.5\%$ and a decrease in final error of $1.84\%$. 
We suspect Action performs better because it directly considers the impact of a message on future behavior, leading to more informed and targeted measurement acquisition at the next step. 
We note that the comparison of relative network loads is more valuable here than the absolute value, as the threshold values used in \Cref{eq:expected-utility} are tunable parameters that should reflect actual network constraints in a real deployment.

\subsection{Oracle Handshaking}
\begin{figure}
         \centering
         \includegraphics[width=\columnwidth]{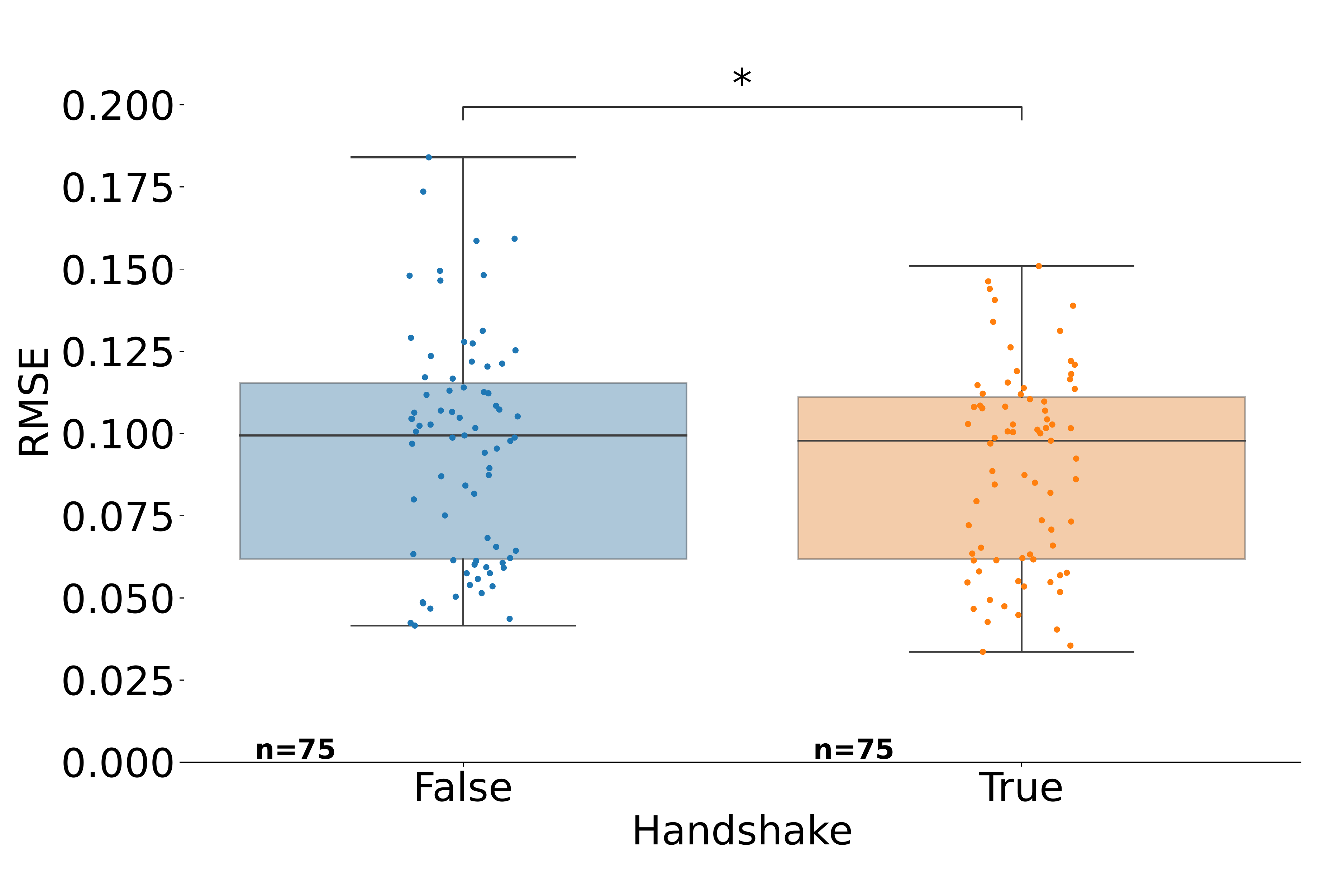}
  \caption{
  Right: the sender receives an oracle handshake
  when a message is successfully transmitted.
  Bar indicates significance under one-sided Wilcoxon signed-rank test; *: $\psig$.
  }
  \label{fig:handshake}
  \vspace{-0.1in}
\end{figure}

We implement a modification to the communication protocol where we assume an oracle handshaking routine, which gives $\ego$ an accurate confirmation handshake from each $\other$ indicating whether the message was successfully received; $\ego$ then only updates $M^{\other}$ when successful.
\Cref{fig:handshake} shows the final RMSE comparing the two; we see that there is a significant improvement with oracle handshaking (Wilcoxon signed-rank test $\wpres{1143}{0.05}$).
Though practically impossible, this suggests that implementing such a (probabilistic) routine in the network could improve performance further; however, the resulting burden of these messages must also be taken into account.

\section{Conclusion}
We presented the first study on restricted communication in the context of quantile estimation for environmental analysis.
We show that restricting message transmission based on whether the actions that the sender believes other robots will take at the next state would change with the message results in a $42\%$ decrease in network load while also decreasing quantile estimation error by $1.84\%$, when compared to transmitting every message possible.
This indicates that targeted communication during quantile estimation  can have real benefits to real-world multirobot deployments when robots and their networks are resource-constrained.

In addition to receipt handshaking, there are several interesting directions of future work.
We believe allowing the sender to choose the most valuable measurements from its collected data rather than restricting it to the most recent measurements may improve performance. 
However, this would lead to challenges in scalability as the environment is explored, and could be addressed by applying a rolling window over the previous measurements to mitigate this.
Further, we are interested in methods for better approximating the models that a robot maintains for each of its teammates, which could be achieved by augmenting message headers with metadata such as that robot's current dataset size or current quantile estimates.

\addtolength{\textheight}{-1cm}   



%
%

\newpage
\bibliographystyle{IEEEtran}
\bibliography{IEEEabrv,references}

\end{document}